\documentclass{article}
\usepackage{amsmath}

\usepackage[preprint]{GuidNav}

\usepackage[utf8]{inputenc} 
\usepackage[T1]{fontenc}    
\usepackage{hyperref}       
\usepackage{url}            
\usepackage{booktabs}       
\usepackage{amsfonts}       
\usepackage{nicefrac}       
\usepackage{microtype}      
\usepackage{xcolor}         
\usepackage{graphicx}

\title{Guiding VLM Agents with Process Rewards at Inference Time for
GUI Navigation}

\author{
 \textbf{Zhiyuan Hu\textsuperscript{1}},
 \textbf{Shiyun Xiong\textsuperscript{1}},
 \textbf{Yifan Zhang \textsuperscript{2}},
 \textbf{See-Kiong Ng\textsuperscript{1}},
\\
 \textbf{Anh Tuan Luu\textsuperscript{3}},
 \textbf{Bo An\textsuperscript{3}},
 \textbf{Shuicheng YAN\textsuperscript{1,2}},
 \textbf{Bryan Hooi \textsuperscript{1}}
\\
\\
 \textsuperscript{1}National University of Singapore,
 \textsuperscript{2}Skywork AI,
 \textsuperscript{3}Nanyang Technological University
\\
}

\begin{document}

\maketitle

\begin{abstract}
 Recent advancements in visual language models (VLMs) have notably enhanced their capabilities in handling complex Graphical User Interface (GUI) interaction tasks. 
Despite these improvements, current frameworks often struggle to generate correct actions in challenging GUI environments.
State-of-the-art commercial VLMs are black-boxes, and fine-tuning open-source VLMs for GUI tasks requires significant resources.
 Additionally, existing trajectory-level evaluation and refinement techniques frequently fall short due to delayed feedback and local optimization issues. 
To address these challenges, we propose an approach that guides VLM agents with process supervision by a reward model during GUI navigation and control at inference time. This guidance allows the VLM agent to optimize actions at each inference step, thereby improving performance in both static and dynamic environments. In particular, our method demonstrates significant performance gains in three GUI navigation tasks, achieving a 3.4\% improvement in single step action accuracy for static environments, along with a around 33\% increase in task success rate in one dynamic environment. With further integration of trajectory reflection and retry mechanisms, we also demonstrate even greater enhancement in task success. 
\end{abstract}

\section{Introduction}

Recent advances in VLMs have significantly enhanced their capabilities in understanding, reasoning, and generalizing, enabling them to handle complex real-world GUI interaction tasks \citep{hong2024cogagent, you2024ferret, cheng2024seeclick}. For instance, given an instruction like ``How do I get to the nearest Walmart?", a VLM agent is expected to navigate to the Google Maps application, search for Walmart locations in the vicinity, and select the nearest one to initiate route navigation. These advancements greatly improve the accessibility and efficiency of GUI interaction tasks.

However, even state-of-the-art visual language models (VLMs) like GPT-4V \citep{openai2023gpt4}, Gemini 1.5 Pro \citep{reid2024gemini} and others, as well as interaction agent frameworks like \cite{yan2023gpt, wang2024mobile, zhang2024android}, still struggle to generate correct actions when completing GUI tasks such as VisualWebarena \citep{koh2024visualwebarena}, OSWorld \citep{xie2024osworld} and others. 
These commercial VLMs are typically black-box models, making them \textbf{inaccessible for tuning}, and further \textbf{fine-tuning open-source VLMs for GUI tasks remains resource-intensive}.
Additionally, \citet{pan2024autonomous} introduce a technique where GPT-4V serves as an evaluator to assess task success and provide reflection for retrying in case of failure, which can enhance the performance of agents in GUI navigation and control. However, such evaluation and refinement methods at the end of a trajectory will lead to \textbf{local optimization deficiency and delayed feedback}. Evaluating only at the end of the trajectory can result in insufficient optimization of individual actions, overlooking the refinement needed at each step. In GUI tasks, where each step impacts the final outcome, neglecting step-by-step optimization may degrade overall performance. Moreover, trajectory-level evaluation delays error correction, increasing both computational and time costs. Meanwhile, \citet{bai2024digirl} propose DigiRL, improving task performance in dynamic environments by combining Advantage-Weighted Regression with online reinforcement learning (RL) and an automatic curriculum mechanism. Such RL methods can lead to \textbf{high computational and time costs, along with a complex training process} that requires extensive online interaction data. Moreover, the training of RL algorithms is often unstable due to the sparsity and uncertainty of feedback, as well as the inherent trade-off between exploration and exploitation. These factors contribute to the high training cost and prolonged training time, particularly in dynamic and complex environments.

\begin{figure*}
    \centering
    \includegraphics[width=0.85\linewidth]{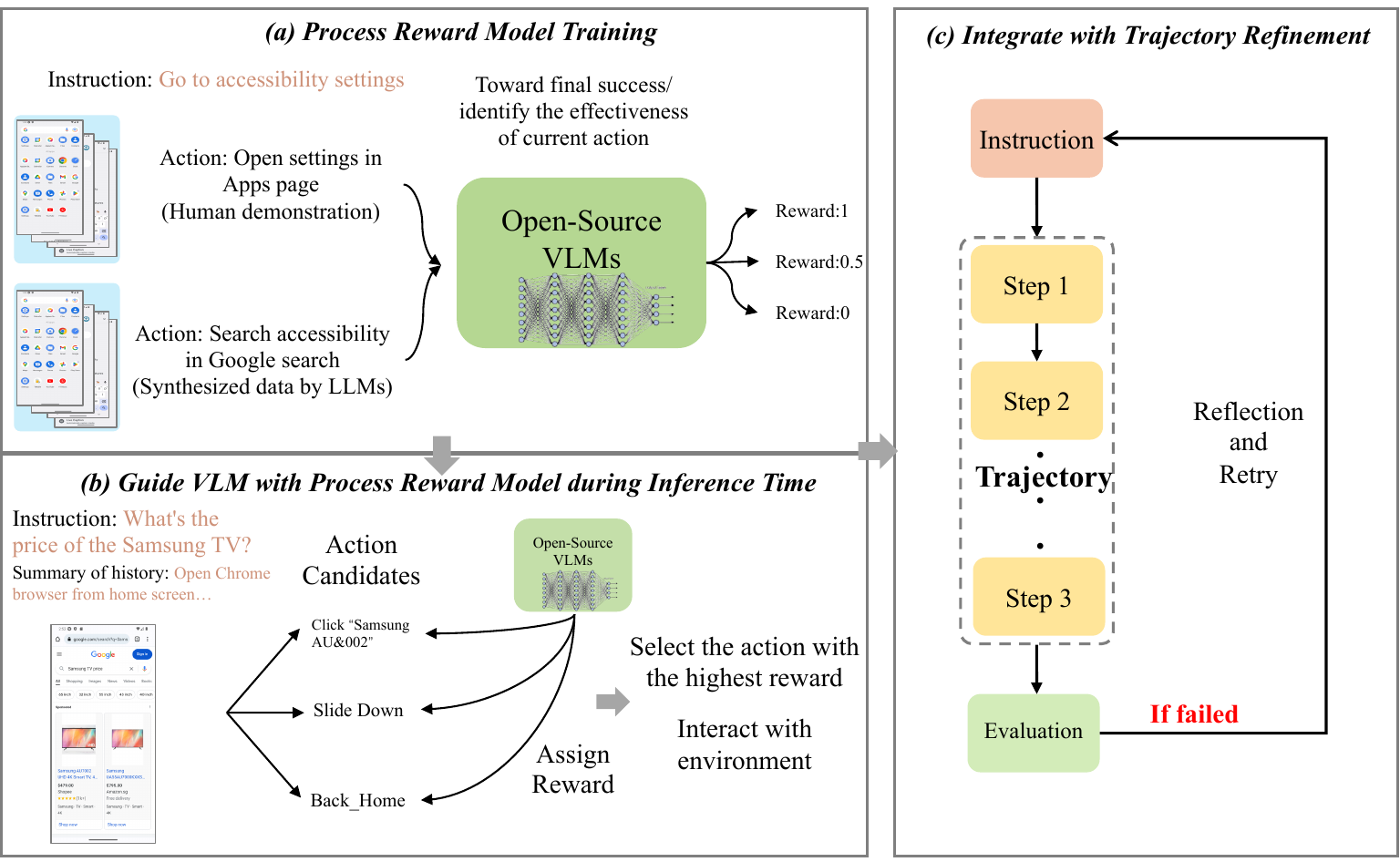}
    \caption{Overview of GuidNav. }
    \label{overview}
\end{figure*}

To address these limitations, we propose \textbf{GuidNav}, guiding the VLM agent with a process reward model during interaction inference in GUI navigation and control tasks. 
Our empirical findings, along with OpenAI's o1 results \citep{openaio12024}, show that increasing effort during inference can significantly improve performance.
Furthermore, there are strong reasons to favor process supervision through a reward model. It offers more precise feedback by identifying the exact step where an error occurs. This approach directly rewards models for following a  path to success. In contrast, models guided by trajectory-based outcome supervision often take inefficient actions, deviating from the correct path and requiring additional effort to correct. Process-based rewards can help reduce these deviations, leading to a more efficient action trajectory. Guiding the VLM agent with this process reward model enables the agent to learn which actions are effective for achieving the given goal within the GUI task environment. 
As illustrated in Figure~\ref{overview}, to achieve this guidance, we first train a process reward model based on a limited amount of human demonstrations and synthesized data generated by the VLM. This process reward model learns the feedback signal from GUI data, guiding the VLM agent during GUI navigation inference to ensure it executes optimal actions. By providing process reward feedback at each inference step, the VLM agent can more accurately adjust its behavior. This fine-grained optimization enhances the success rate of tasks, especially in complex GUI environments where the correct execution of each step is crucial. Furthermore, unlike delayed feedback in trajectory-level evaluation, step-level process rewards enable the model to learn and adapt to environmental changes in real-time, preventing the accumulation of errors caused by delayed feedback. Additionally, like the demonstration in Figure~\ref{overview} (c), our process reward model can also been integrated into outcome supervision pipline to further improve action generation and selection.

We evaluate GuidNav in both static and dynamic settings within Android-in-the-Wild (AitW) \citep{rawles2024androidinthewild}, measuring action accuracy at each step based on existing annotations and overall task success through human evaluation. The experimental results demonstrate that our method improves GPT-4o by approximately 5\% in action accuracy of static environments and about 33\% in task success rate within dynamic environments. With trajectory reflection and retry mechanisms, the success rate can reach a peak of 71.6\%. We also urther evaluate GuidNav on two additional benchmarks, GUI Odyssey \citep{lu2024gui} and Mind2Web \citep{deng2023mind2web}, to demonstrate its generalization ability. GuidNav achieves an average improvement of approximately 3.2\% across three LLMs on GUI Odyssey, and a 2.1\% increase on Mind2Web, both under static environments for single-step action accuracy. In summary, our contributions are primarily in the following three areas:

\begin{itemize}
    \item We introduce GuidNav, an approach that guides VLM agents for action decision during GUI interactions via a process reward model.
    \item Our approach can be easily integrated into trajectory-level refinement to further strengthen the performance.
    \item We demonstrate that our method enhances VLM agents in both static and dynamic settings, and we further validate its general effectiveness across three benchmarks.
\end{itemize}

\section{Related Work}
\textbf{GUI Navigation Agents and Benchmarks}
Previous GUI navigation agents primarily focused on text-based prompts describing the environment, such as the HTML code, Document Object Model (DOM), or accessibility trees. However, current research leverages both screenshots and text instructions to navigate interfaces more akin to human-environment interactions. For instance, Auto-UI \citep{zhan2023you} utilizes GUI data to tune the language and projection modules, enabling interaction in multimodal GUI environments without the need for environment parsing or application-dependent API access. AppAgent \citep{yang2023appagent} employs the vision capabilities of large language models to operate smartphone applications in a human-like manner. Mobile-Agent-v2 \citep{wang2024mobile} presents a multi-agent architecture for assisting mobile device operations. OS-Copilot \citep{wu2024copilot} accelerates the development of computer agents on Linux and MacOS by providing a universal interaction interface. MM-Navigator \citep{yan2023gpt} generates executable actions based on the screen image, text instructions, and interaction history. CogAgent \citep{hong2024cogagent} leverages extensive GUI grounding data to further train the VLM for enhanced interaction. Additionally, several works focus on visual interaction tasks across app, web, and OS environments. AitW \citep{rawles2024androidinthewild} and Weblinx \citep{lu2024weblinx} use human demonstrations to evaluate the accuracy of proposed actions. Osworld \citep{xie2024osworld}, AgentStudio \citep{zheng2024agentstudio}, AndroidWorld \citep{rawles2024androidworld}, and Visualwebarena \citep{koh2024visualwebarena} provide simulation environments for executing arbitrary agent trajectories in various domains and tasks. While these are not yet perfect, they serve as suitable platforms for assessing agents' capabilities.

\textbf{Evaluation by Reward and Reinforcement Learning Methods} 
In addition to the agent framework, some researchers use reinforcement learning and reward models to enhance VLM agents further. \citet{pan2024autonomous} introduce an Autonomous Evaluator for agent behavior, refining the agent’s ability through reflection or fine-tuning based on filtered behavior cloning data. \citet{bai2024digirl} improve task performance in dynamic environments by combining Advantage-Weighted Regression with online reinforcement learning and an automatic curriculum mechanism. \citet{zhai2024fine} employ a reinforcement learning method with a game-rule-based reward to strengthen VLM-powered agents in Gym Cards and ALFWorld. \citet{fereidouni2024search} utilize a two-stage learning process: Supervised Learning, where human demonstration data maps state to action, and Unsupervised Learning, where the PPO algorithm fine-tunes by optimizing policy gradients. A language model calculates action probabilities based on user goals and observations.
Compared to process reward models, these methods often struggle to provide fine-grained step-level feedback and tend to incur significantly higher training costs.

\textbf{Discussion} Unlike methods like DigiRL \citep{bai2024digirl} and Autonomous Evaluator \citep{pan2024autonomous} that rely on trajectory-level feedback or high computational training costs, GuidNav provides step-level rewards, enabling immediate optimization and reducing computational costs. This approach offers more efficient and precise action refinement, improving task performance in both static and dynamic environments.

\section{Method}

The primary task of GUI navigation involves enabling the VLM agent to interpret task instructions and interact with GUI screenshots to achieve a desired goal. Formally, let $x$ represent the task instruction, and let $S_t = \{s_1, s_2, \dots, s_t\}$ represent the sequence of GUI states observed at different time steps $t$. The VLM agent must generate and select actions $A_t = \{a_1, a_2, \dots, a_t\}$, where $a_t$ is the action taken at time step $t$, that modify the GUI state from $s_t$ to $s_{t+1}$. The overall goal is to generate and determine an action sequence $A$ that leads the GUI environment from the initial state $s_1$ to a final state $s_T$ that satisfies the task's objective. The core challenge is to enable the VLM agent to generate and determine actions $a_t$ at each time step that are most aligned with the task's goal $g$, while minimizing errors or irrelevant operations. The method, GuidNav, we propose consists of a two-stage process aimed at training the process reward model and guide  VLMs agent for efficient task execution in GUI task environments at each time step $t$. The stages are as follows:

\subsection{Reward Model Training}
As shown in Figure~\ref{overview} (a) and task definition, given a user instruction $x$, the historical states $S_{t-1} = \{s_1, s_2, \dots, s_{t-1}\}$ and corresponding actions $A_{t-1} = \{a_1, a_2, \dots, a_{t-1}\}$, as well as the current state $s_t$ and an action candidates $a_t$, the VLM as process reward model $\mathbb{R}$ assigns a reward $r$ to a given action candidate $a_t$ in the context of the user instruction $x$ and current $(S_{t-1},A_{t-1})$. The training data for this process reward model can be obtained through two primary sources:

\begin{enumerate}
    \item \textbf{Human Demonstrations}: This involves collecting data from human experts who interact with the environment, providing a trajectory of states, actions, and corresponding rewards, i.e., $\{x^{(i)}, (S^{(i)}, A^{(i)}, R^{(i)})\}_{i=1}^N$, where $R^{(i)}$ represents the reward sequence for trajectory $i$. The reward value $r_{t}$ at time step $t$ of a trajectory is typically set to 1, as each action candidate is carefully selected by human experts and is assumed to be correct by default.

    \item \textbf{Self-Playing via VLMs}: VLMs simulate interactions with the environment, generating synthetic trajectories of states and actions $\{x^{(j)}, (S^{(j)}, A^{(j)}, \tilde{R}^{(j)})\}_{j=1}^M$, where $\tilde{R}^{(j)}$ denotes the synthesized reward sequence. The reward $r_{t}$ is assigned based on the effectiveness of the VLM-generated action candidate $a_t$ in achieving the user instruction $x^{(j)}$ and the task's goals, rather than merely its similarity to human demonstrations. Details regarding the value assignment of $r_t$ are provided in Appendix~\ref{reward}.

\end{enumerate}

To reduce the input length, thus mitigating potential degradation in performance due to excessively long inputs, at each time step, the VLM is employed to perform multimodal self-summarization based on the prompt $P$ (elaborated in Appendix~\ref{prompts}), which converts the historical state and actions $(S,A)$ into a concise history in the form of natural language: 
\begin{equation}
    h_t = \textbf{VLM}((S_{t-1}, A_{t-1}), P))
\end{equation}

Then, the reward $r_t$ guided by process reward model $\mathbb{R}$ and assigned to action candidate $a_t$  can be represented as:
\begin{equation}
    r_t = \mathbb{R} (x, h_t, s_t, a_t)
    \label{reward_model}
\end{equation}

\textbf{Training Objective} \quad The process reward model $\mathbb{R}$ is trained to minimize the difference between the predicted rewards and the annotated rewards by minimizing a loss function. Specifically, the objective is to minimize the Mean Squared Error across all trajectories and their respective time steps between the predicted reward $r_{t,\text{pred}}$ for each action candidate and the annotated reward $r_{t, \text{anno}}$:
\begin{equation} \mathcal{L}(\theta) = \frac{1}{\sum_{i=1}^{N} T^{(i)}} \sum_{i=1}^{N} \sum_{t=1}^{T^{(i)}} \left( r_{t, \text{pred}}^{(i)} - r_{t, \text{anno}}^{(i)} \right)^2 \end{equation}

Here, $N$ represents the total number of trajectories, and $T^{(i)}$ is the number of time steps in trajectory $i$. The term $r_t^{(i)}$ refers to the predicted reward at time step $t$ in the $i$-th trajectory, and $r_{t, \text{true}}^{(i)}$ is the corresponding annotated reward.

\subsection{Guide VLMs with a Process Reward Model}
We demonstrate the process in Figure~\ref{overview} (b), as the aforementioned reward model training.

\textbf{Action Generation} \quad We follow the similar strategy for VLM interaction inference in GUI tasks. Given a user instruction $x$, the historical states $S_{t-1} = \{s_1, s_2, \dots, s_{t-1}\}$ and corresponding actions $A_{t-1} = \{a_1, a_2, \dots, a_{t-1}\}$, the VLM serves as policy model $\mathbb{P}$ will first summarize the previous states and actions to obtain a concise history summary $h_t$. Thus, the user instruction $x$, history summary $h_t$, current time step state $s_t$ and corresponding prompt $P_\text{inference}$ (Appendix~\ref{prompts}) will be used as input of VLM to generate $k$ possible actions $\mathcal{A}_t = {a_t^1, a_t^2, \dots, a_t^k}$. This can be formulated as:
\begin{equation}
\mathcal{A}_t = \mathbb{P} (x, h_t, s_t, P_\text{inference})
\end{equation}

\textbf{Reward Assignment} \quad According to Equation~\ref{reward_model}, the reward model assigns a scalar reward $r_t^k$ for each action candidate $a_t^k$ based on its alignment with the task. The reward $r_t^k$ is calculated as:
\begin{equation}
    r_t^k = \mathbb{R} (x, h_t, a_t^k, s_t)
\end{equation}
    
\textbf{Action Selection} \quad The VLM selects the action $a^*$ with the highest reward $r^*$:
\begin{equation} 
    a_t^* = \arg\max_{a_t^k} r_t^k 
\end{equation}
    
This selected action $a^*$ is then executed to interact with the environment. The process is iteratively refined to improve the alignment of the VLM's actions with the desired outcomes, ensuring that the actions taken are those most likely to achieve the user's objective.

\subsection{Trajectory Refinement and Evaluation}
\label{Integration}
The process reward model can also been integrated with the refinement of the trajectories generated by the VLM to further enhance the performance, as depicted in Figure~\ref{overview} (c).
\paragraph{\textbf{Trajectory Formation}} \quad Once an action $a^*$ is selected and executed, it becomes part of the trajectory $T = {(s_1, a_1), (s_2, a_2), \dots, (s_t, a_t)}$. The trajectory is the sequence of state-action pairs from the initial state to the task objective.

\paragraph{\textbf{Evaluation and Reflection}} \quad At the end of each trajectory, the VLM evaluates the success of the trajectory in achieving the desired outcome. If the trajectory fails to meet the desired criteria, the VLM reflects on the reasons for the failure, generating a ``reflection thought" that encapsulates the lessons learned from the unsuccessful attempt. This reflective thought is then incorporated into the retry process, informing the next iteration.
\paragraph{\textbf{Reflection and Retry}} \quad The reflective thought generated by the VLM becomes part of the input for the next attempt after the whole trajectory. The VLM uses this enriched input, including the original instruction and the new reflective thought, to generate a new trajectory. This iterative process of reflection and retry continues until the VLM successfully achieves the task objective. Once a successful trajectory is identified, it is confirmed, and the process is completed.

\section{Experiment}

\subsection{Baselines}

\textbf{Direct Prompting (DP)} involves directly prompting a Visual Language Model (VLM) to generate an action based on the instruction query, the current screenshot, and a summary of the previous state.
\vspace{1mm}

\textbf{TopK} is a technique where the model generates the top $k$ most probable actions \citep{xiong2023can,tian2023just}. In this procedure, while the model generates $k$ actions (with $k$ set to 3 in our work), we simplify the process by always selecting the most probable one (the first action in the list). This ensures that the model still considers multiple possibilities but prioritizes the highest-probability action for execution.
\vspace{1mm}

\textbf{Reflection} \citep{shinn2024reflexion}  is a framework that improves LLMs' decision-making abilities in various tasks by using linguistic feedback and episodic memory, achieving significant performance gains in environments.
\vspace{1mm}

\textbf{Autonomous Refinement (AR)} \citep{pan2024autonomous} leverages the Reflexion
technique \citep{shinn2024reflexion}, an agent first attempts a task, and an external evaluator is used to judge whether its attempt was successful or not. If unsuccessful, the agent reflects and retries, guided by GPT-4o as evaluator.
\vspace{1mm}

\textbf{DigiRL} \citep{bai2024digirl} is an autonomous RL approach,  for training in-the-wild device control agents through fine-tuning a pre-trained VLM in two stages: offline RL to initialize the model, followed by offline-to-online RL.

\subsection{Dataset}
To evaluate the effectiveness and generalization ability of our approach across different platforms and interaction modalities, we conduct experiments on three benchmarks: AitW \citep{rawles2024androidinthewild}, GUI Odyssey \citep{lu2024gui}, and MIND2WEB \citep{deng2023mind2web}. These datasets cover mobile GUI navigation and open-world web navigation tasks, offering diverse and challenging environments. For all datasets, the full action spaces are provided in Appendix~\ref{action space}.

\vspace{1mm}
\textbf{AitW} is a large-scale dataset for Android device control, comprising 715,142 human demonstrations across 30,378 unique instructions. These instructions are divided into four subsets: General, WebShopping, GoogleApps, and Installation. We leverage ground-truth data and self-play data from 300 tasks in each subset to form the training set for the process reward model. Following the approach outlined in \cite{yan2023gpt}, we randomly select 300 tasks from the AitW test set to evaluate action accuracy in a static environment. To ensure balanced representation, each subset contributes 75 tasks. For dynamic environment evaluation, we sample 120 tasks (30 from each subset) as instruction queries in a simulated setting.

\vspace{1mm}
\textbf{GUI Odyssey} 
 is a comprehensive dataset for cross-application GUI navigation on mobile devices, consisting of 7,735 navigation episodes collected across 201 mobile applications. The tasks span six categories, including General Tools, Information Management, Web Shopping, Media Entertainment, Social Sharing, and Multi-Apps, and typically involve multi-step workflows that require transitioning between multiple apps. Following the evaluation settings described in the original paper, we randomly sample 500 tasks from the GUI Odyssey test set to construct our evaluation set. To ensure robust assessment, we conduct experiments across three different policy models. Action accuracy is measured by comparing the predicted actions with ground-truth annotations at each step. 

\vspace{1mm}
\textbf{Mind2Web}
 is a large-scale dataset for web navigation, consisting of 2,350 human demonstrations across 137 real-world websites and 31 domains. Each task involves completing multi-step operations such as clicking, typing, and selecting. Due to the diversity and complexity of tasks, Mind2Web serves as a challenging benchmark for evaluating the generalization ability of our method in web environments. For evaluation, 80 tasks are randomly sampled from each of the Cross-Task, Cross-Website, and Cross-Domain subsets. Action accuracy is measured based on the alignment between predicted and ground-truth action sequences.  

\subsection{Setup}

We use diverse VLMs (e.g., GPT-4o \citep{hurst2024gpt}, Gemini 2.0 Flash \citep{gemini2flash}, and Qwen-VL-Plus \citep{ali}) as the VLM policy model and CogVLM2 \citep{hong2024cogvlm2} as the process reward model. The Set-of-Mark (SoM) method \citep{yang2023set} facilitates VLM understanding and interaction with the screen by providing discrete tags for UI elements, enabling direct selection without requiring precise coordinate localization (see Appendix~\ref{SoM}). Experiments run with batch size 4, learning rate 1e-4, AdamW, cosine decay, 0.05 warmup ratio, and 2 epochs.

In \textbf{static evaluation}, we assess the correctness of model actions at each step across three datasets: AitW, GUI Odyssey, and Mind2Web. For AitW and GUI Odyssey, we compute the screen-wise partial action matching score, where given the historical states, current state, and user instructions, the model’s predicted action is compared against the ground-truth action following the protocols in prior work \citep{rawles2024androidinthewild, yan2023gpt}. The static evaluation metric is defined as the number of correct actions divided by the total number of steps per episode. For Mind2Web, static evaluation instead measures the element selection accuracy: a step is correct if the selected UI element matches any acceptable target defined by the benchmark. The primary goal of static evaluation across all datasets is to ensure that the model generates appropriate actions at each step. In \textbf{dynamic assessment}, we focus on overall task performance rather than stepwise correctness. Two human annotators evaluate whether each task is successfully completed, emphasizing the model’s ability to achieve the final goal in an open-ended, uncertain environment.

\begin{table}[ht]
\centering
\small

{
\setlength{\tabcolsep}{3pt} 

\begin{tabular}{l|ccccc}

\toprule
{\bf Method} & General  & GA & Install & WS & Average \\
\midrule 
VLM-FT300 & 28.9  & 34.9 & 37.1 & 33.2 & 32.3 \\
VLM-FT800 & 45.1 & 44.8 & 54.6 & 45.5 & 47.9 \\
\bottomrule
\end{tabular}
}

\caption{Performance comparison of different numbers of generated trajectories data (GA represents Google Apps and WS is Web Shopping). VLM-FT300 refers to the model fine-tuned with 300 successful trajectories, and VLM-FT800 indicates fine-tuning with 800 successful trajectories. We evaluate in a static environment with corresponding metrics, using 75 tasks per subset.}
\label{self-improvement}
\end{table}

\begin{figure}
    \centering
    \includegraphics[width=0.8\linewidth]{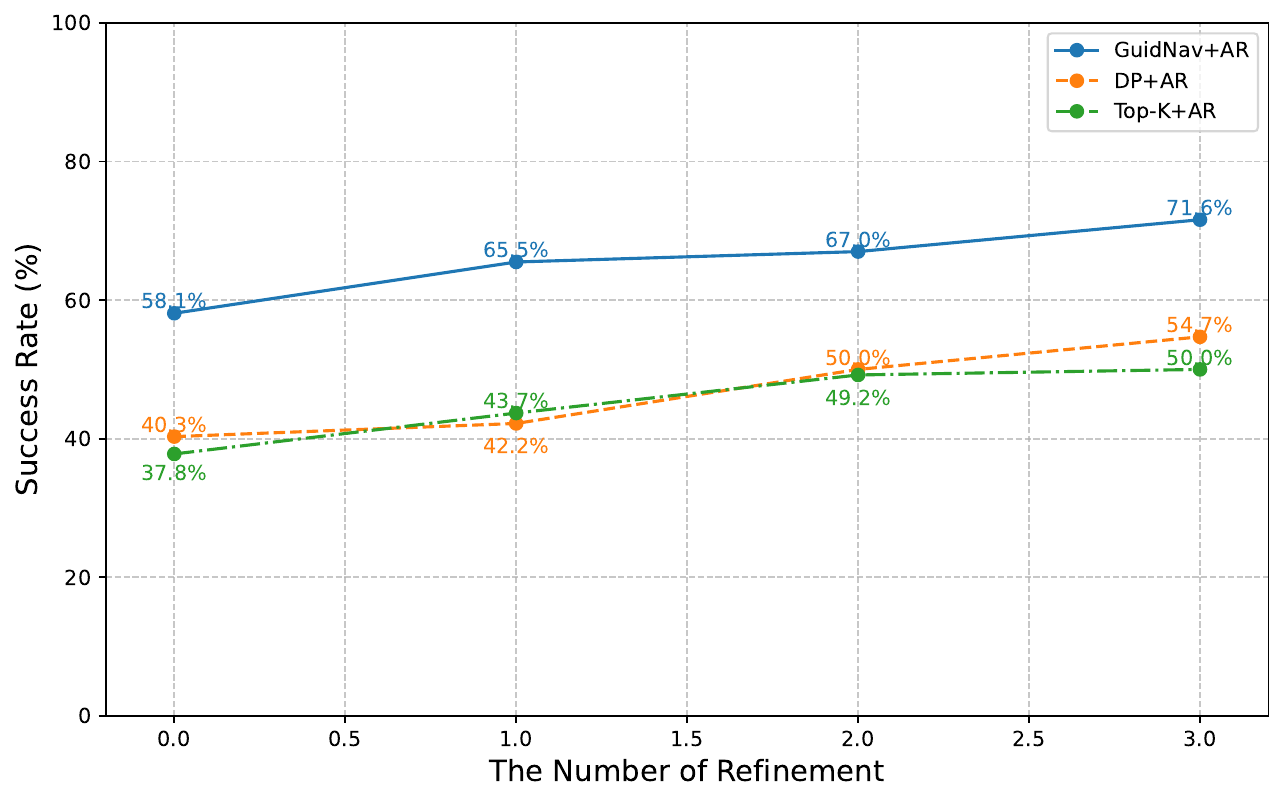}
    \caption{The performance curve across different trial numbers shows the impact of refinement techniques. `DP+AR' represents the combination of direct prompting for action at each step, followed by AR at the end of each trajectory trial. `GuidNav+AR' integrates TopK action selection guided by a reward model, with AR applied at the end of each trajectory trial. `TopK+AP' refers to TopK method integrated with AR.}
    \label{integration}
\end{figure}

\section{Performance}

\begin{table*}[h]
\centering
{
\begin{tabular}{l|ccccc}
\toprule
& General & Google\_apps & Install & Web\_shopping & Average \\
\midrule
Topk w/ Oracle Eval & 55.8 & 49.5 & 54.4 & 57.2 & 53.7 \\
\midrule
DP & 30.3 & 39.1 & 36.2 & 34.4 & 34.3 \\
TopK & 31.0 & 35.8 & 34.4 & 36.9 & 34.0 \\
Reflection \citep{shinn2024reflexion} & 31.2 & 37.9 & 32.6 & 30.0 & 32.9 \\

\midrule
GuidNav & 35.5 & 41.4 & 40.9 & 38.5 & 38.9 \\
GuidNav Pass@N & \textbf{43.4} &\textbf{48.4} &\textbf{48.8} &\textbf{42.3} & \textbf{46.8} \\
\bottomrule
\end{tabular}
}
\caption{Performance comparison of approaches in \textit{static assessment} across four AitW tasks. Topk w/ Oracle Eval uses an oracle to select the best action from the top-K candidates. Pass@N, with N set to 3, calculates the action accuracy across multiple attempts, counting how many outcomes are correct in 3 trials.}
\label{AITW static}
\vspace{-3mm}
\end{table*}

\begin{table*}[h]
\centering
{
\begin{tabular}{l|ccccc}
\toprule
& General & Google\_apps & Install & Web\_shopping & Average \\
\midrule
DP & 52.9 & 48.4 & 33.3 &20.8  &40.3 \\
Topk &47.1 &45.2 & 40.0 &12.5  &37.8 \\
AR \citep{pan2024autonomous} & 58.8 & 38.7 & 53.3 & 16.7 & 43.7 \\
DigiRL \citep{bai2024digirl} & 56.3  & - & - & 32.7 &-  \\
GuidNav & 64.7 & 71.0 &60.7 &29.2 &58.1 \\
Integration & \textbf{82.4} & \textbf{83.9} & \textbf{75.0} & \textbf{34.8} &\textbf{71.6}  \\
\bottomrule
\end{tabular}
}
\caption{Performance comparison of approaches in \textit{dynamic assessment} across four AitW tasks. Integration refers to the method where we combine the process reward model with the AR approach (3 retries), as detailed in Section~\ref{Integration}.}
\vspace{-2mm}
\label{AITW dynamic}
\end{table*}
\subsection{Experimental performance in AitW}
\textbf{Static Evaluation} \quad
As shown in Table~\ref{AITW static}, compared to DP, our method achieves an average improvement of 4.6\%, particularly in the `General' domain tasks where we see a 5.2\% gain. Simply applying the TopK method does not yield the same benefit. However, when we incorporate a reward model to identify the most likely action, the improvement becomes even more substantial. Furthermore, among the top $k$ possible actions, selecting the action based on oracle evaluation (Topk w/ Oracle Eval) reveals a high upper bound. This indicates that while VLMs can generate a potentially correct action, the correct one is often not their first choice. Applying our method multiple times (GuidNav Pass@N) boosts performance significantly.

\textbf{Dynamic Assessment} \quad
In the dynamic environments of table~\ref{AITW dynamic}, our GuidNav outperforms both DP and AR, achieving overall improvements of approximately 17.8\% and 14.4\%, respectively. Even when compared to the DigiRL method, which includes further tuning via reinforcement learning, GuidNav maintains superiority. Additionally, our method provides process-level supervision, whereas the AR approach evaluates the final outcome and offers insights for retries. These two methods can be naturally integrated (the results of `Integration'), enabling us to achieve a higher success rate with a maximum of three retries. We also report the results based on Gemini 2.0 Flash to add more comparable baselines in Appendix~\ref{Gemini baseline}.

\textbf{Accuracy of Step-wise Reward Model}. \quad We evaluate the step-wise reward model on 500 AitW steps, achieving 78.8\% accuracy. The model determines correctness at each step using ground truth and random actions as inputs, alongside historical state, current state, and user instructions.

\subsection{Experimental analysis in GUI Odyssey}
Table~\ref{gui odyssey performance} presents a performance comparison of different models on the GUI Odyssey dataset  under static environment settings. Following the evaluation metric used in the original benchmark, we report results based on the same metric definitions. Across all models, the inclusion of our reward model consistently led to marked improvements. For instance, \textit{GPT-4o} showed a 3.2\% absolute gain, \textit{Qwen-VL-Plus} improved by 5.0\%, and \textit{Gemini-2.0-flash} improved by 2.1\%. These gains translate into relative improvements of 14.1\%, 28.2\%, and 7.9\%, respectively. Compared to the Top-K with Oracle Evaluation, there still remains a gap. However, the application of the reward model significantly narrows the performance difference between standard inference and oracle action selection. These results strongly affirm the effectiveness and broad applicability of our method across different architectures and complex, real-world navigation environments.
\begin{table*}[h]
\centering
\begin{tabular}{lccc}
\toprule
\textbf{} & \textbf{GPT-4o} & \textbf{Gemini-2.0-Flash} & \textbf{Qwen-VL-Plus} \\
\midrule
Topk w/ Oracle Eval  & 36.0 & 36.3 & 31.0 \\
\midrule
Without Reward Model  & 22.7 & 26.7 & 17.7 \\
With Reward Model & 25.9 & 28.8 & 22.7 \\
\bottomrule
\end{tabular}
\caption{ Performance comparison of policy models on the GUI Odyssey test set. Topk w/ Oracle Eval uses an oracle to select the best action from the top-K candidates.}
\label{gui odyssey performance}
\vspace{-4mm}
\end{table*}

\begin{table*}[h]
\centering
\small
\begin{tabular}{l|cc|cc|cc|cc}
\toprule
& \multicolumn{2}{c|}{Cross-Task} & \multicolumn{2}{c|}{Cross-Website} & \multicolumn{2}{c|}{Cross-Domain} & \multicolumn{2}{c}{Overall} \\
& Ele. Acc & Step SR & Ele. Acc & Step SR & Ele. Acc & Step SR & Ele. Acc & Step SR  \\
\midrule
Topk w/ Oracle Eval & 58.8 & 54.5 & 73.6 & 66.5 & 63.5 & 60.3 & 65.4 & 60.4 \\
\midrule
Without reward model & 46.9 & 43.2 & 54.8 & 50.7 & 51.1 & 47.6 & 50.9 & 47.1 \\
With reward model & 48.2 & 44.5 & 58.8 & 54.8 & 51.6 & 47.9 & 53.0 & 49.2 \\
\bottomrule
\end{tabular}
\caption{Performance comparison of approaches in static assessment on the Mind2Web dataset across Cross-Task, Cross-Website, and Cross-Domain settings. Topk w/ Oracle Eval uses an oracle to select the best action among the top-K candidates. Models are evaluated with and without a reward model, measuring Element Accuracy (Ele. Acc) and Step Success Rate (Step SR).}
\label{Mind2Web performance}
\vspace{-5mm}
\end{table*}
\subsection{Experimental results in Mind2Web}
We evaluate our method on the Mind2Web dataset, following the Min2Web benchmark evaluation metric. \textit{Ele. Acc}  measures whether the selected element matches any acceptable target, and \textit{Step SR} requires both the element and the operation to be correct. Each step is evaluated independently based on the ground-truth action history.
Table~\ref{Mind2Web performance} shows the comparison results. Under static environment settings, without applying the process reward model, the baseline achieves an overall Element Accuracy of 50.9\% and a Step Success Rate of 47.1\%. After introducing the reward model, the performance improves to 53.0\% and 49.2\%, respectively, corresponding to a 2.1\% improvement on both metrics. Moreover, across different generalization settings, our method consistently improves performance. On the \textit{Cross-Task} split, it brings an absolute improvement of 1.3\% in both Element Accuracy and Step Success Rate. On the \textit{Cross-Website} split, larger gains are observed, with Element Accuracy increasing by 4.0\% and Step Success Rate by 4.1\%. On the \textit{Cross-Domain} split, the improvement in Element Accuracy is 0.5\%, while Step Success Rate improves slightly by 0.3\%. 

Overall, the results validate that our framework not only generalizes well in Android environments, but also enhances performance and generalization when applied to complex real-world web tasks as evaluated on Mind2Web.
\subsection{Self-improvement in Open Source VLM Policy Model} 

Based on the guidance provided by the process reward model, the policy model can be enhanced during inference. Specifically, for the results presented in Table~\ref{self-improvement}, we collect additional data via self-play on the AitW dataset, where GPT-4o generates top-k candidate actions at each decision step in a static environment. These candidates are then evaluated against the ground truth to determine the quality of each action, and the resulting feedback is used to refine the reward model. The model's intrinsic abilities are strengthened as a result of the process reward model's guidance through successful trajectory fine-tuning. Table~\ref{self-improvement} shows the benefit of increasing successful trajectories. With 300 or 800 trajectories, model's performance improves further.

\subsection{Overall Performance of GuidNav} 

We evaluated our method, GuidNav, across three challenging datasets: AitW, GUI Odyssey, and Mind2Web. Averaging across these datasets, our approach yielded an overall 3.4\% improvement in static environments. Furthermore, in dynamic environments, GuidNav achieved substantial gains. notably improving task success rates by approximately 33\%. GuidNav enhances both step-level action accuracy and task completion in dynamic, uncertain environments. Its consistent improvements across mobile and web GUI tasks highlight its generalization to diverse interaction settings.

\section{Analysis}

We analyze our method on AitW, focusing on AR integration, efficiency, and case-based effectiveness.

\subsection{Integration with Autonomous Refinement}

As mentioned earlier, our method can also be integrated with an AR approach. Specifically, we continue to guide the VLM agent with the reward model at each step and use GPT-4o for reflective feedback. We run up to three rounds to evaluate AR and integration.

As shown in Figure~\ref{integration}, the performance of the AR method significantly improves from the first to the third round, though the incremental benefit decreases with each additional round. However, the performance curve of the integrated method with the process reward consistently remains above that of the AR (DP+AR) alone, indicating our approach consistently improves AR effectiveness.

\subsection{Comparison of Computational Efficiency}

\begin{figure*}[t]
    \centering
    \includegraphics[width=0.7\linewidth]{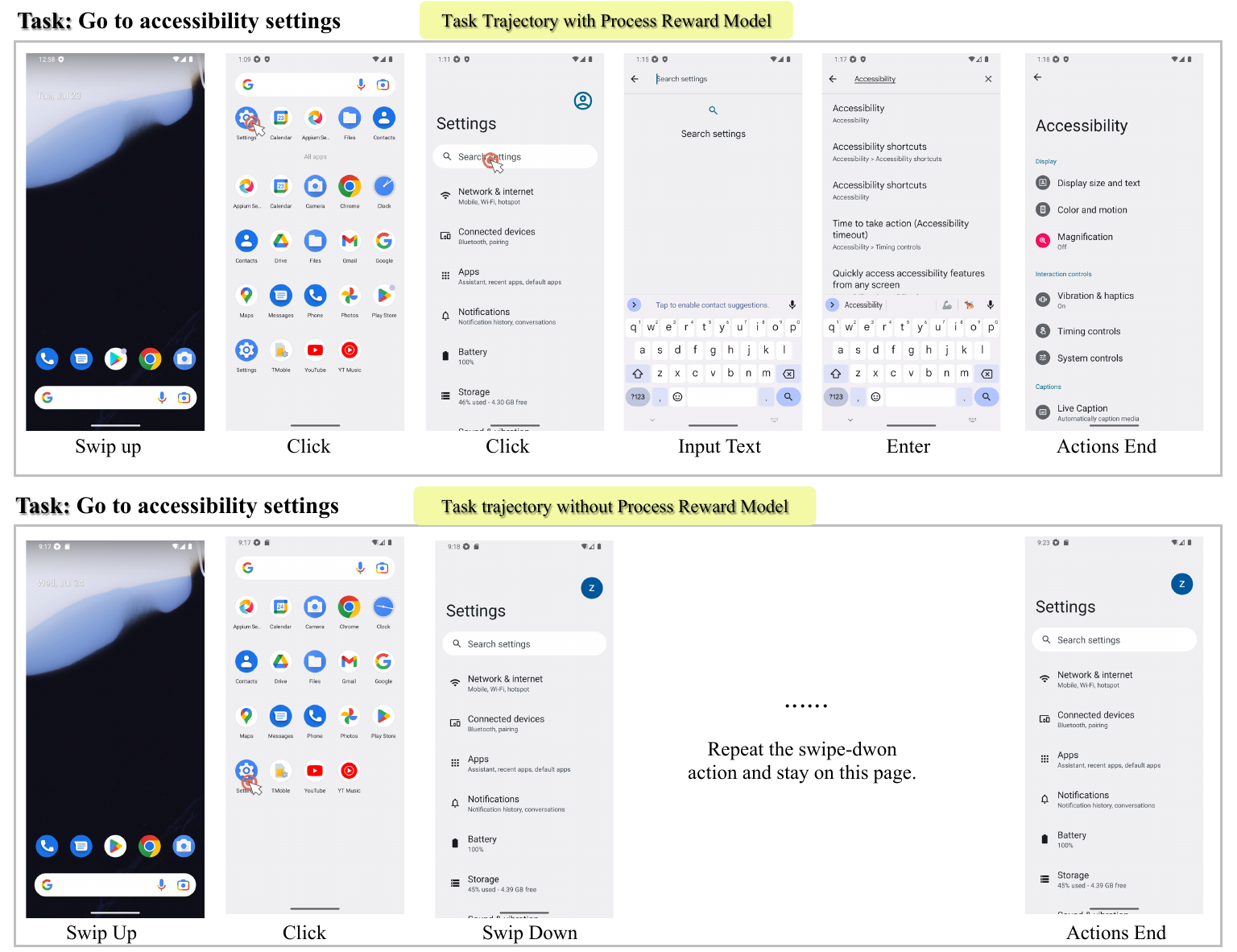}
   \vspace{-3mm}
    \caption{Example of case study. Access the accessibility settings.}
    \label{case2}
\end{figure*}

We compare the efficiency of several methods across different metrics. Table~\ref{result: token cost} presents the average token consumption, API cost, and average number of interaction turn required to complete a task for each method. 
First, the DP and TopK methods show noticeable differences in token consumption and cost. DP consumes 41.0k tokens with a cost of \$0.13, while TopK consumes 59.4k tokens, resulting in a higher cost of \$0.23. Additionally, the number of interaction turns is lower for DP (8.7 turns) compared to TopK (10.6 turns).
The AR (n=1) and Integration (n=1) methods consume more tokens, 102.1k and 108.6k respectively, resulting in higher costs of \$0.40 and \$0.41. 
However, AR (n=2) consumes the most tokens, 129.4k, leading to the highest cost of \$0.51, showing a significant increase in cost with the number of evaluation rounds. These results suggest that the DP and GuidNav methods are more cost-effective, especially for tasks requiring more interaction turn. Although AR methods consume more tokens and incur higher costs, they may offer advantages like improved accuracy or other performance benefits, depending on task requirements.

\begin{table}[ht]
\centering
\scalebox{0.85}{ 
\setlength{\tabcolsep}{8pt} 
\begin{tabular}{l|ccc}
\toprule
{\bf Method} & Tokens  & Cost & Turn \\
\midrule 
DP & 41.0k  &0.13    & 8.7 \\
TopK & 59.4k  &0.23  &10.6    \\
GuidNav & 53.4k  &0.20 &9.8 \\
AR(n=1) & 102.1k  &0.40    &17.8  \\
AR(n=2) & 129.4k  &0.51   & 22.5 \\
Integration(n=1) & 108.6k  &0.41   & 19.8 \\
\bottomrule
\end{tabular}
}
\caption{Efficiency measurement for different methods is analyzed across several metrics. ``Tokens" refers to the average token consumption per task. The ``Cost" metric corresponds to the average API cost per task, based on the latest GPT-4o pricing (\$5.00 per 1M tokens). ``Turn" indicates the average number of interaction turn. The variable $n$ denotes the number of evaluation rounds.}
\label{result: token cost}
\end{table}

\subsection{Case Study}

Our method demonstrates a superior ability to select appropriate actions for complex tasks, particularly those requiring precise operation sequences. As shown in Figure \ref{case1}, when searching for flights in Google Chrome, the correct procedure involves clicking the search bar before typing, as highlighted in the previous action summary. Our method correctly follows this sequence, ensuring accurate and efficient interaction. In contrast, GPT-4o types the search content without clicking the search bar first, leading to an incorrect operation.

Moreover, our approach outperforms others in efficiency. Our reward model learns to prioritize actions that advance the task goal with minimal redundancy. This is evident in how our method focuses on relevant actions, avoiding unnecessary exploration. For instance, in the bottom scenario of Figure \ref{case2}, when accessing accessibility settings, other methods like VLM lack precision, repeatedly swiping down and getting stuck in a blind search. In contrast, our method swiftly identifies the correct path, avoiding redundant actions and enabling more reliable task execution in dynamic environments.

\section{Conclusion}

In this work, we presented the approach to guiding VLMs with a process reward model for improved performance in GUI interaction tasks. Our method addresses the limitations of existing frameworks by enabling VLM agents to optimize actions at each inference step, significantly enhancing action accuracy and task success rates in both static and dynamic environments. Specifically, we demonstrate a near 5\% improvement in action accuracy in static GUI environments and a around 15\% increase in task success rate in dynamic settings. These results highlight the effectiveness of our process reward model guidance strategy in overcoming challenges such as delayed feedback and local optimization. By integrating trajectory reflection and retry mechanisms, we further improve the success rate and efficiency of VLM agents in complex GUI navigation tasks. For future work, GuidNav’s generalization can be tested beyond specific apps using benchmarks that span operating systems, professional tools, and workflows. Expanding evaluations in these areas will provide deeper insights into its effectiveness in diverse real-world contexts.

\section{Limitations and Future Works}

The current evaluation of GuidNav has primarily focused on specific applications and narrow task domains, which, while demonstrating its potential, leaves room for further exploration of its capabilities in more diverse and complex real-world scenarios. Future work should aim to rigorously test GuidNav’s generalization across a broader spectrum of environments, including cross-platform compatibility (e.g., Windows, macOS, Linux, mobile OS), specialized professional tools (e.g., CAD software, IDEs, data analysis platforms), and dynamic, multi-step workflows. By developing comprehensive benchmarks and conducting extensive evaluations in these areas, we can better assess GuidNav’s scalability, adaptability, and robustness, ultimately unlocking its full potential to handle a wide range of tasks and challenges in real-world settings. This expansion will not only validate its effectiveness but also push the boundaries of task automation in more complex and varied contexts.

\newpage
\bibliographystyle{ACM-Reference-Format}
\bibliography{GuidNav}





\newpage
\appendix
\section{Appendix}
\subsection{Reward Annotation}
\label{reward}

   To collect training data for the reward model, we utilize three datasets: AitW~\citep{rawles2024androidinthewild}, GUI Odyssey~\citep{lu2024gui}, and Mind2Web~\citep{deng2023mind2web}. For each dataset, we employ a self-play strategy in a static environment, using GPT-4o or the corresponding policy model to generate actions. We automatically label each action at the step level by evaluating its effectiveness. The general labeling rule is consistent across datasets: a predicted action is considered correct if both the action type and the operation or gesture match the ground truth. For AitW, since the ground truth is represented as coordinates while our model outputs numeric labels for detected elements, we treat any point within the detected bounding elements as equivalent. For GUI Odyssey and Mind2Web, the evaluation is based on matching the selected element and operation type according to their respective benchmark protocols. In all cases, evaluation metrics are aligned with the original definitions of each dataset.
    \begin{itemize}
    \item \textbf{For click actions}: The action is considered correct if the chosen element by action is within 14\% of the screen distance from the ground truth coordinate, or if both the ground truth coordinate and the element selected by generated action fall within the same detected bounding box (expanded to 240\% of its original size for action matching).
    \item \textbf{For scroll actions}: the predicted action is considered correct if the scroll direction (up, down, left, or right) matches the ground truth direction.
    \item \textbf{For other actions}: For other actions: The predicted action is considered correct if the action type matches the ground truth. However, for the typing action, both the action type and the typed content must match the ground truth.
    \end{itemize}
\subsection{SoM Settings}
\label{SoM}

    SoM utilizes off-the-shelf interactive segmentation models, such as SEEM \citep{zou2024segment} or SAM \citep{kirillov2023segment}, to partition an image into regions of varying granularity. Each region is annotated with marks like alphanumeric labels, masks, or bounding boxes. This enhances the VLM's ability to interpret and understand image elements.
    
    In our implementation, we maintain the original SoM configuration (input an image, output bounding boxes and corresponding labels, and overlay the labels onto the image). We choose SAM as the segmentation model and, for each identified entity, assign a unique numeric label positioned at the center. Additionally, we store the labeled screenshot along with the coordinates of each labeled entity for subsequent interactions.
    
    Given the complex structure of GUI interfaces, which often include numerous small entities with relationships such as containment and overlap, we implement specific strategies to ensure accurate interpretation by the VLM. For entities in containment relationships, we retain the identifiers of both the containing and contained entities. In cases of overlapping entities, we prioritize the identifier of the smaller entity to ensure clarity and precision.
\subsection{Action Space of all Datasets}
\label{action space}

The AitW dataset contains predefined actions for VLM agents in Android GUI navigation tasks. The actions are represented as follows:
\begin{itemize}
    \item \texttt{click}: Perform a click action on a UI element with a specific \texttt{id}. Example: \{\texttt{action\_type}: \texttt{click}, \texttt{id}: \texttt{<numeric IDs on the screen>}\}
    \item \texttt{type}: Input text into a UI element. Example: \{\texttt{action\_type}: \texttt{type}, \texttt{text}: \texttt{<text>}\}
    \item \texttt{navigate\_home}: Navigate back to the home screen. Example: \{\texttt{action\_type}: \texttt{navigate\_home}\}
    \item \texttt{navigate\_back}: Navigate to the previous screen. Example: \{\texttt{action\_type}: \texttt{navigate\_back}\}
    \item \texttt{enter}: Confirm the current action, typically mimicking an 'enter' key press. Example: \{\texttt{action\_type}: \texttt{enter}\}
    \item \texttt{scroll}: Scroll in a specified direction (\texttt{up}, \texttt{down}, \texttt{left}, \texttt{right}). Example: \{\texttt{action\_type}: \texttt{scroll}, \texttt{direction}: \texttt{up}\}
    \item \texttt{task\_complete}: Mark the task as completed. Example: \{\texttt{action\_type}: \texttt{task\_complete}\}
\end{itemize}

The predefined actions of the GUI Odyssey dataset are represented as follows:
\begin{itemize}
    \item \texttt{click}: Perform a click action on a UI element with a specific \texttt{id}. Example: \{\texttt{action\_type}: \texttt{click}, \texttt{id}: \texttt{<numeric IDs on the screen>}\}
      \item \texttt{longpress}: Perform a longpress action on a UI element with a specific \texttt{id}. Example: \{\texttt{action\_type}: \texttt{longpress}, \texttt{id}: \texttt{<numeric IDs on the screen>}\}
    \item \texttt{type}: Input text into a UI element. Example: \{\texttt{action\_type}: \texttt{type}, \texttt{text}: \texttt{<text>}\}
    \item \texttt{navigate\_home}: Navigate back to the home screen. Example: \{\texttt{action\_type}: \texttt{navigate\_home}\}
    \item \texttt{navigate\_back}: Navigate to the previous screen. Example: \{\texttt{action\_type}: \texttt{navigate\_back}\}
    \item \texttt{scroll}: Scroll in a specified direction (\texttt{up}, \texttt{down}, \texttt{left}, or \texttt{right}). Example: \{\texttt{action\_type}: \texttt{scroll}, \texttt{direction}: \texttt{up}\}
\end{itemize}

The Mind2Web dataset contains predefined actions for VLM agents in Web navigation tasks, which are represented as follows:
\begin{itemize}
    \item \texttt{click}: Perform a click action on a UI element with a specific \texttt{id}. Example: \{\texttt{action\_type}: \texttt{click}, \texttt{id}: \texttt{<numeric IDs on the screen>}\}
    \item \texttt{type}: Input text into a UI element. Example: \{\texttt{action\_type}: \texttt{type}, \texttt{text}: \texttt{<text>}\}
\end{itemize}

\subsection{Gemini 2.0 Flash Performance}
\label{Gemini baseline}
\begin{table}[ht]
\centering
\small
    {
    \setlength{\tabcolsep}{3pt} 
    \begin{tabular}{l|ccccc}
        \toprule
         & General & Google\_apps & Install & Web\_shopping & Average \\
        \midrule 
        Topk &38.2 &19.4 & 10.5 &20.8  &24.1 \\
        AR & 55.9 & 29.0 & 27.8 & 29.2 & 37.4 \\
        \bottomrule
    \end{tabular}
    }
\caption{Performance comparison of approaches in dynamic assessment across four AitW tasks, using Gemini 2.0 Flash as the policy model. The results for the TopK and AR \citep{pan2024autonomous} methods, when evaluated without the reward model guidance, demonstrate inferior performance compared to our approach.}
\end{table}

\subsection{Prompts}

\label{prompts}
\begin{table*}[ht]
    \centering
    \begin{tabular}{p{\linewidth}}
    \hline
    \textbf{Instruction:} \\
    \hline
    Provide a summary of the previous actions as follows:
    \{previous\_text\}
    , the current thinking steps and the action to be executed as follows:
    \{text\}, and the screenshot of the interface after the action is executed.
    Please summarize the actions above and the status after the action is executed into the new previous actions using descriptive languages brief as possible.(do not speculate on the next move )
    Summary: \\
    \bottomrule
    \end{tabular}
     \caption{Prompt for generating historical summarization}
\end{table*}

\begin{table*}[!t]
    \centering
    \begin{tabular}{p{\linewidth}} 
    \toprule
    \textbf{Task:} Goal of task \\
    \midrule
    \textbf{Task Requirements:} \\
    \midrule
    Above are two screenshots of a android phone. one is the original screen and the other one has blocks with numeric IDs. You are an AI assistant with a deep understanding of these screenshot and the android phone operations. \\For example, The home page does not display all installed apps, scrolling up on the home page can open the App Drawer where all the installed apps [If an app is not in the app drawer, it is not installed.] are stored and organized, or you can check whether an app is installed in Google Shop. You need to generate an action based on the current situation, which will be executed automatically without user intervention. \\The user will not interfere with the entire operation process, such as voice input, which will be regarded as an incorrect operation. \\ Attention: When the user can find the answer from the current page (without needing detailed information), the task can be considered complete."\\
    \midrule
    \textbf{Available Actions:} \\
    \midrule
    {available\_actions} \\
    \midrule
    \textbf{Summary of previous actions:} \\
    \midrule
    \textit{Previous actions}: {previous\_actions} \\
    \midrule
    \textbf{Instruction:} \\
    \midrule
    Based on the above information and the following instruction. please provide your \textit{k} best thought processes (think step by step) and answers for the next one action(only one action),then provide the probability (0.0 to 1.0) that each action contributes to completing the user's requirement at the current stage (according to the image). \\
    Answer format for example:\\
    G1: \textless the step-by-step explanation of your thought process (No more than three sentences)\textgreater\ So the next one action is:\{"action\_type": \textless action type in \textless Available Actions\textgreater, \textless the rest information of the action\textgreater\} \\
    P1: \textless the probability between 0.0 and 1.0 that G1 is correct, without any extra commentary whatsoever; just the probability!\textgreater\\
    ...\\
    Gk: \textless the step-by-step explanation of your thought process (No more than three sentences)\textgreater\ So the next one action is:\{"action\_type": \textless action type in \textless Available Actions\textgreater, \textless the rest information of the action\textgreater\} \\
    Pk: \textless the probability between 0.0 and 1.0 that Gk is correct, without any extra commentary whatsoever; just the probability!\textgreater 
    \\
    \bottomrule
    
    \end{tabular}
   \caption{Prompt for generating $k$ possible actions} 
\end{table*}

\end{document}